\documentclass{ws-procs11x85}
\usepackage{ws-procs-thm}           
\usepackage[table]{xcolor} 

\usepackage{longtable}
\usepackage{multirow}
\usepackage{array}
\usepackage{multirow}
\usepackage{tikz}
\usepackage{fontawesome5}
\usepackage{colortbl}      
\newcolumntype{P}[1]{>{\raggedright\arraybackslash}p{#1}}

\usepackage{soul}
\definecolor{myRed}{RGB}{255,220,220}
\definecolor{myGreen}{RGB}{220,255,220}
\definecolor{lightGray}{RGB}{245,245,245} 
\newcommand{\hlred}[1]{{\sethlcolor{myRed}\hl{#1}}}
\newcommand{\hlgreen}[1]{{\sethlcolor{myGreen}\hl{#1}}}
\usepackage{booktabs}
\definecolor{lightgray}{gray}{0.9}
\definecolor{headercolor}{RGB}{222,235,247} 
\definecolor{rowcolor}{RGB}{239,243,251}    

\begin{document}
\title{ReXErr: Synthesizing Clinically Meaningful \sethlcolor{myRed}\hl{Errors} in \\ Diagnostic Radiology Reports}
\author{Vishwanatha M. Rao$^\dag$$^1$, Serena Zhang$^\dag$$^1$, Julian N. Acosta$^1$, Subathra Adithan$^2$, Pranav Rajpurkar$^1$}

\address{$^1$Department of Biomedical Informatics, Harvard Medical School
Boston, MA 02115, USA\\
$^2$Department of Radiodiagnosis, Jawaharlal Institute of Postgraduate Medical Education and Research, India \\
E-mail: : vishwanatha.rao@pennmedicine.upenn.edu, serena2z@stanford.edu, julian\_acosta@hms.harvard.edu, subathra.a@jipmer.edu.in}

\begin{abstract}
Accurately interpreting medical images and writing radiology reports is a critical but challenging task in healthcare. Both human-written and AI-generated reports can contain errors, ranging from clinical inaccuracies to linguistic mistakes. To address this, we introduce ReXErr, a methodology that leverages Large Language Models to generate representative errors within chest X-ray reports. Working with board-certified radiologists, we developed error categories that capture common mistakes in both human and AI-generated reports. Our approach uses a novel sampling scheme to inject diverse errors while maintaining clinical plausibility. ReXErr demonstrates consistency across error categories and produces errors that closely mimic those found in real-world scenarios. This method has the potential to aid in the development and evaluation of report correction algorithms, potentially enhancing the quality and reliability of radiology reporting.
\end{abstract}


\copyrightinfo{$^\dag$Authors contributed equally to this work.}


\section{Introduction}\label{aba:sec1}

Radiology reports provide crucial information for clinical decision-making and patient outcomes\cite{cote_forecasting_2018}. However, creating radiology reports is an intensive process, and requires a trained specialist to analyze medical images and write in-depth medical reports\cite{reiner_radiology_2007,al_yassin_it_2018}. In human-written reports, errors can arise due to various factors such as fatigue, high case volumes or complexity. These errors may include misinterpretation of imaging findings, incomplete documentation of relevant clinical information, and inconsistencies in terminology and language usage. In addition to such inaccuracies, the subjective nature of radiological interpretation leaves room for errors, which may go unnoticed until they impact patient care\cite{bruno_understanding_2015,brady_error_2017}. 

Recently, there has been a significant push towards automating the creation of these reports using deep learning. While current approaches to generating radiology reports have, in some cases, succeeded in creating complete and clinically relevant reports\cite{zhou_generalist_2024,tu_towards_2024,wu_towards_2023,tanida_interactive_2023}, automated report generation presents its own set of challenges stemming from inherent biases within algorithms, model constraints, and limitations in the data used. Errors can range from references to non-existing priors, which are easier to detect, to false predictions or omissions, which are much more problematic clinically and often go unnoticed\cite{messina_survey_2022,sloan_automated_2024}. The prevalence of errors, both in radiologist-written as well as AI-based reports, leaves a great need for more comprehensive tools that can screen for and correct them. Throughout this paper, we present the Chest X-Ray Report Error (ReXErr) method that can generate errors at a report and sentence level. ReXErr offers a novel pipeline to synthesize plausible errors that capture the breadth and diversity of errors made by humans and models and can thus be used to generate data to train and adapt error correction algorithms. Figure 1 outlines an overview of the error generation process.

\begin{figure}[t!]
\centerline{\includegraphics[width=\linewidth]{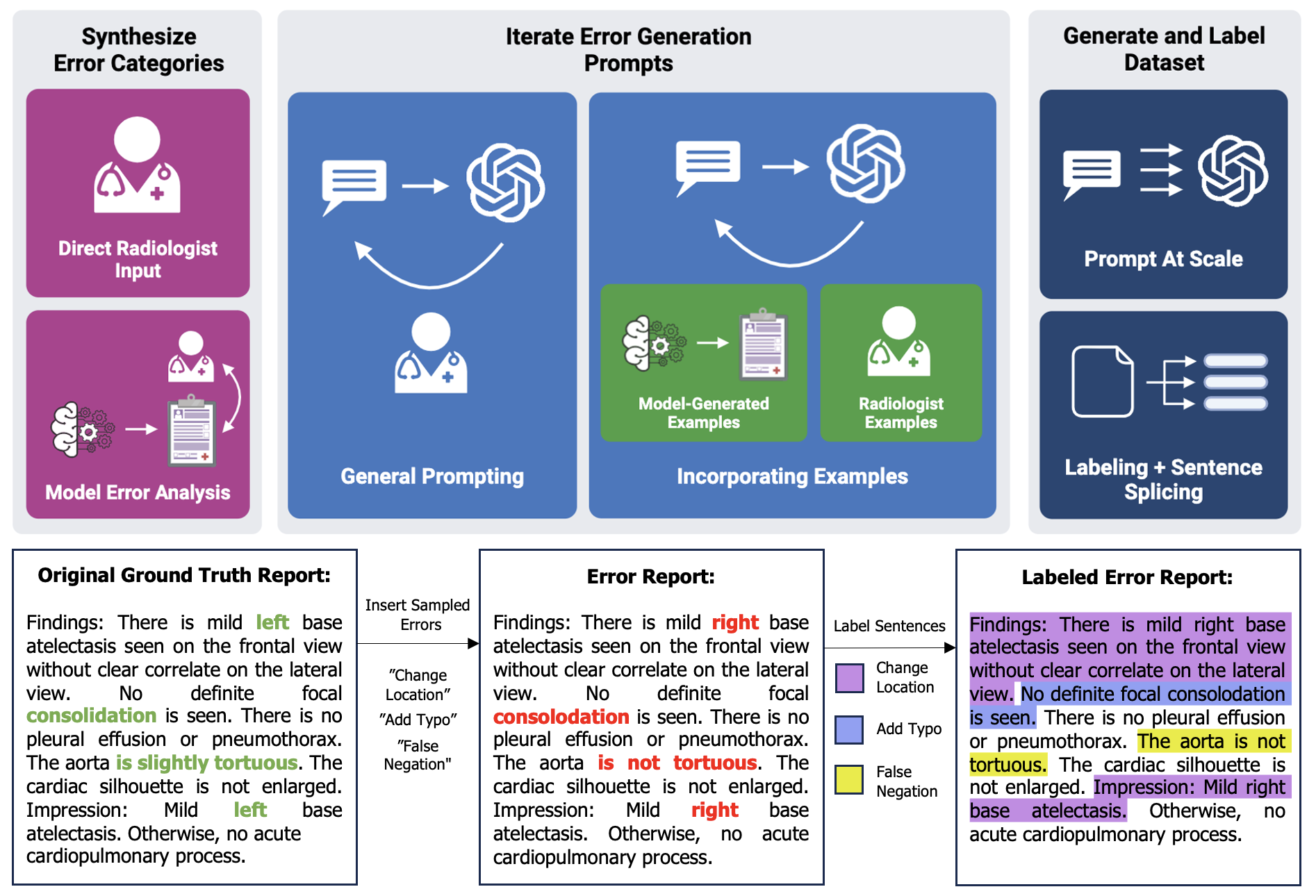}}
\caption{Summary of ReXErr error generation pipeline. The bottom panel provides an example of applying ReXErr to a sample radiology report.}
\label{aba:fig1}
\end{figure}
\vspace{-10pt}

\section{Related Work}\label{aba:sec1}
\subsection{Error Classification in Radiology Reports}
While prior work has attempted to categorize the types and causes of mistakes made by clinicians, errors introduced through report generation models fall in a different distribution and require their own dedicated analysis. One example is a categorization framework developed by Yu et al. to analyze common errors in model-generated radiology reports, aiming to create metrics that account for these errors and improve alignment with clinician feedback\cite{yu_evaluating_2023}. Their framework includes six categories: “False prediction of finding”, “Omission of finding”, “Incorrect location/position
of finding”, “Incorrect severity of finding”, “Mention of comparison that is not present in
the reference impression”, and “Omission of comparison describing a change from a previous
study.” Others have created datasets to categorize the errors commonly made in retrieval-based report generation models by their severity level and then correct each error using either deletion, substitution, or insertion of a line\cite{tian_refisco_nodate}. Another metric, FineRadScore, was developed to identify line-by-line corrections required to go from the candidate to the ground-truth report. This metric categorizes each line based on the type and severity of each correction, using the six categories for type proposed by Yu et al. and the categories for severity defined by Tian et al\cite{huang_fineradscore_2024}. 

\begin{table}[t!]
\tbl{Summary of the errors incorporated within the ReXErr pipeline.}{%
\arrayrulecolor[HTML]{AAACED}
\resizebox{\textwidth}{!}{%
\fontsize{8}{10}\selectfont
{\fontfamily{pcr}
\begin{tabular}{>{\raggedright\arraybackslash}p{4cm}>{\raggedright\arraybackslash}p{4cm}>{\raggedright\arraybackslash}p{5.5cm}}
\hline
\textbf{Error Type} & \textbf{Error Category} & \textbf{Specific Errors} \\
\hline
\multirow{11}{4cm}{AI Generated Report Errors} & \multirow{3}{3.5cm}{Content Addition} & Add Medical Device \\
 & & False Prediction \\
 & & False Negation \\
\cline{2-3}
 & \multirow{2}{3.5cm}{Linguistic Quality} & Add Repetitions \\
 & & Add Contradictions \\
\cline{2-3}
 & \multirow{6}{3.5cm}{Context-Dependent} & Change Name of Device \\
 & & Change Position of Device \\
 & & Change Severity \\
 & & Change Location \\
 & & Change Measurement \\
\hline
\multirow{3}{4cm}{Human Errors} & Content Addition and Context-Dependent & Human error - similar to above \\
\cline{2-3}
 & \multirow{2}{3.5cm}{Linguistic Quality} & Change to Homophone \\
 & & Add Typo \\
\hline
\end{tabular}
}}}
\end{table}

\subsection{Error Detection}
The earliest works for error detection in radiology reports have focused on specific issues like laterality errors, gender mismatches, or single-word errors, often using simple matching techniques rather than deep contextual analysis\cite{lee_detection_2015,minn_improving_2015}. Subsequent research developed more sophisticated methods using LSTM architectures for error detection\cite{zech_detecting_2019}. One study created an error generation system using CheXpert labels and constructed a more comprehensive error detection approach using a BERT-based model\cite{min_rred_2022}. More recently, GPT-4 has been shown to identify common errors in radiology reports with a level of proficiency comparable to the average radiologist\cite{gertz_potential_2024}. However, this study was limited to five error categories: omission, insertion, spelling, side confusion, and a general ``other" category.
\subsection{Report Correction}
Efforts have also been made to develop models to correct a specific type of hallucination in AI-generated radiology reports: false references to non-existent prior scans or reports. One study adapted two different models, BioBERT and GPT-3, to rewrite the impression sections of radiology reports without references to prior scans\cite{ramesh_improving_2022}. Another study applied direct preference optimization on medical vision language models to suppress unwanted references to priors\cite{banerjee_direct_2024}. While these approaches address a crucial aspect of report accuracy, they focus on a narrow subset of potential errors in AI-generated reports.

Report revision is an emerging task that focuses on refining and correcting existing radiology reports. Unlike report generation, which creates reports from a scratch, revision aims to improve the accuracy, completeness, and clarity of pre-existing reports through an instruction prompt detailing what to correct for certain sentences in the report.  This task has gained importance as AI systems increasingly assist in radiology workflows, necessitating methods to verify and enhance AI-generated or human-authored reports. 

Recent research has integrated report revision capabilities into larger, multi-functional foundation models. Zhao et al. introduced a practical report generation dataset and method that incorporates various clinical interactions and contextual information\cite{zhao_large_nodate}. This benchmark includes a dedicated instruction dataset specifically designed for report revision. Other foundation models have been designed with interactive dialogue components that potentially allow for report rewriting and revision through an instruction prompt\cite{zhou_generalist_2024,tu_towards_2024,wu_towards_2023,yang_customizing_2023,moor_med-flamingo_2023,zhao_chatcad_2023,xu_elixr_2023}. However, while report revision addresses the refinement of existing reports given an explicit instruction, it is unable to identify and rectify underlying errors or omissions in the absence of clear directives. This limitation underscores the need for complementary approaches that can not only revise reports but also detect and correct inaccuracies autonomously, thereby enhancing the overall quality and reliability of radiological reporting. 

Interestingly, there is a lack of studies exploring the utility of error-filled report versions without accompanying correction instructions, which presents an opportunity for further research. Such error-filled reports could serve valuable purposes: they could be used as negative examples in reinforcement learning algorithms to improve AI model performance, or as a means to validate automatic evaluation metrics like RadCliQ. These potential applications highlight the broader implications of developing a robust error injection method and dataset for radiology reports.

\begin{table}[t!]
\tbl{Baseline prompting description for each error category.}{%
\arrayrulecolor[HTML]{AAACED}
\resizebox{\textwidth}{!}{%
\fontsize{8}{10}\selectfont
{\fontfamily{pcr}
\begin{tabular}{>{\raggedright\arraybackslash}p{4.2cm}>{\raggedright\arraybackslash}p{11.8cm}}
\hline
\textbf{Error} & \textbf{Baseline Instruction / Description} \\
\hline
Add Medical Device & Add sentences that could be part of a radiology report regarding the presence of one or more devices such as these: pacemaker, central venous line, NG tube, ET tube, ICD. \\
\hline
Change Name of Device & If there is a medical device present in the report, change the name of the medical instrument to a different name that is clinically plausible. \\
\hline
Change Position of Device & If there is a medical device location present in the report, change the position of the medical instrument to a different position that is clinically plausible. \\
\hline
Change Severity & Change the severity of a finding in the report in a manner that makes clinical sense (e.g., change `mild' to `moderate'). \\
\hline
Change Location & Change the location or anatomy of a finding in the report in a manner that is still clinically accurate (e.g., change `right' to `left' or `lateral' to `medial'; always modifying a sentence). \\
\hline
False Prediction & Add a finding that is not present in the report (either adding a sentence or modifying a sentence to insert). \\
\hline
False Negation & Change a particular finding from the report from present to absent by changing a sentence to indicate absence of the positive finding. \\
\hline
Change Measurement & If there is a measurement for a device/finding present, change the units of measurement (e.g., change `cm' to `mm') or change the value of the measurement to a different but still reasonable value (e.g. change `4.9 cm' to `5.8 cm'). \\
\hline
Add Opposite Sentence & Add/alter a statement that is the opposite of another statement earlier in the same report. \\
\hline
Add Repetitions & Add repetitions of sentences present within the report. \\
\hline
Change to Homophone & Change a word in the report to a homophone of that word. \\
\hline
Add Typo & Add a typographical error in the report. \\
\hline
\end{tabular}
}}}
\end{table}

\subsection{Synthetic Data for Report Generation}
Despite their promise, few studies have utilized generative models to create new pipelines for dataset synthesis. Some have attempted to generate image datasets using adversarial learning techniques such as GAN and diffusion models\cite{teixeira_generating_2018,waheed_covidgan_2020,tang_disentangled_2021,liu_utilizing_2023,kashima_cascaded_2023}, while comparatively fewer have used LLMs to generate textual CXR Report datasets, with one using GPT to paraphrase reports present within the MIMIC dataset to extend their training set\cite{hyland_maira-1_2023}. Therefore, there is a need for further studies exploring the use of synthetic textual data in developing report generation models.

\section{Methods}
We created a streamlined pipeline to inject errors into radiology reports, which can be used downstream to generate large datasets and train models for the identification and revision of incorrect radiology reports. We demonstrate error generation with the ReXErr pipeline using reports from the MIMIC-CXR train, dev, and test sets\cite{johnson_mimic-cxr_2019}. This pipeline supports two main tasks: report correction and sentence-level entailment. For both tasks, sentences are classified into three categories: correct (0), error (1), and neutral (2). Neutral sentences reference past reports, findings, or scans and are categorized separately, as algorithms would not be able to determine their accuracy without additional context.

\textbf{Report correction:} Our pipeline generates paired ground truth and error reports, with each error report containing three errors sampled from 12 possible error categories. To highlight the locations of error insertions, we include a dictionary mapping sentence indices from the error report to corresponding indices in the original report. This dictionary indicates the type of error inserted as well as the category of each sentence within the error report. We also separately specify the three error categories used in generating each report.

\textbf{Sentence-level entailment:} We provide a separate pipeline to create a sentence-level error categorization by splicing pairs of sentences from ground truth and error reports. Each pair includes the original sentence and its error version, their label (see categories below), type of error injected (error class) and sequence in the original report (index). Maintaining the sequential detail can help sentence-level entailment models developed upon data generated through CXR-REGen use contextual information from previous sentences to identify errors such as repetitions and contradictions.
\subsection{Error Categories}

Three board-certified radiologists were consulted in synthesizing the final list of errors included within this generation protocol. The errors fall under two broad categories: AI generated report errors and human errors. We further identify three sub-categories of errors: content addition, context-dependent, and linguistic quality errors. Each of the 12 final error categories fall under one of these subcategories and one of the two broader categories. The particular errors were determined in careful collaboration with radiologists; specifically, we used a set of radiologist-annotated reports generated from a current state of the art model\cite{zhou_generalist_2024} to determine the most salient automated generation errors, and consulted radiologists directly to gain a sense for human errors. Table 1 contains a summary of the errors implemented.

\subsection{Data Synthesis}
After extensive iteration and feedback from clinical experts, we developed a comprehensive pipeline for introducing plausible errors into radiology reports using GPT-4o\cite{openai_team_hello_2024}. We define ``plausible" errors as those that either a human or an AI model could realistically make. The pipeline employs a sophisticated sampling strategy to inject errors across all three categories within each report. Context-dependent errors are only introduced when the associated context is present, as determined by regex-based labeling that searches for specific keywords in each report. For instance, errors related to changing the location and type of medical devices are only injected if a device is mentioned in the report. Our approach balances the need for diverse and plausible errors while maintaining the overall structure and believability of the reports. The problem formulation for the injection of errors across all three categories is represented in Equation 1, where $E_C$, $E_A$, and $E_L$ represent context-dependent, content addition, and linguistic quality errors respectively. $T$ refers to the tags present, where $T \in \{\text{``device"}, \text{``measurement"}, \text{``location"}, \text{``severity"}\}$.
\begin{equation}
P(E_c, E_A, E_L \mid T) = P(E_c \mid T) \times P(E_A) \times P(E_L)
\label{aba:eq1}
\end{equation}
The probability of selecting both the content addition ($E_A$) and  linguistic quality ($E_L$) errors are shown below in Equation 2. $A$ and $L$ both represent the number of individual errors present within the content addition and linguistic quality error categories respectively across both the AI and human groups. In our case, $A$ would be 3 and $L$ would be 4, where $L$ includes the linguistic quality errors in both the AI and human error categories.
\begin{align}
P(E_A) &= \frac{1}{|A|} & P(E_L) &= \frac{1}{|L|}
\end{align}
The probability of selecting a context-dependent error given a particular tag is given by Equation 3 below, where the error for the context-dependent error category is sampled across the other categories if no relevant context is present. In the case where multiple tags are provided, the probability of selecting a particular context-dependent error given a tag depends on the normalized weight assigned to the tag ($w'(t_i)$) as well as the total number of context-dependent errors associated with each tag ($E(t)$).
\\
\begin{equation}
P(E_c \mid T = t_i) = 
\begin{cases} 
\frac{w'(t_i)}{\sum_{t \in T} w'(t) \times E(t)} & \text{if } T \neq \emptyset \\
\frac{1}{|A| + |L|} & \text{if } T = \emptyset 
\end{cases}
\end{equation}
\vspace{10pt}
The weights assigned to each tag $w(t_i)$ was calculated based on the frequency of each tag present within the reports through the equations shown below. Each weight is equal to the inverse of the prevalence of its respective tag. The weights are then normalized to $w'(t)$.
\begin{align}
w(t) &= \frac{1}{f(t)} & W &= \sum_{t \in T} w(t) & w'(t) &= \frac{w(t)}{W}
\end{align}
Based on this sampling strategy, GPT-4o was then used to inject the errors. Table 2 summarizes the baseline instructions given for each error type. Appendix A contains the complete long-form prompt used to prompt GPT, whereas Appendix B contains the particular prompts for each error category, including the examples for the relevant errors that use them. 

\begin{table}[t!]
\tbl{Examples of ground truth and error report generated through the ReXErr pipeline.}{%
\arrayrulecolor[HTML]{AAACED}
\resizebox{\textwidth}{!}{%
\fontsize{8}{10}\selectfont
{\fontfamily{pcr}
\begin{tabular}{>{\raggedright\arraybackslash}p{6.25cm}>{\raggedright\arraybackslash}p{6.25cm}>{\raggedright\arraybackslash}p{3.0cm}}
\hline
\textbf{Ground Truth} & \textbf{Error Report} & \textbf{Errors Injected} \\
\hline
\textbf{Findings:} Findings: The patient is status post median sternotomy and CABG. \hlgreen{The heart size is top normal.} The mediastinal and hilar contours are unremarkable. \hlgreen{Bilateral calcified pleural plaques are seen diffusely} which limits assessment of the underlying pulmonary parenchyma. No focal consolidation, pleural effusion or pneumothorax is clearly demonstrated. There are no acute osseous abnormalities. \vspace{+5pt}

\textbf{Impression:} \hlgreen{Bilateral calcified pleural plaques} indicative of prior asbestos exposure.  No definite acute cardiopulmonary abnormality otherwise noted.
 & \textbf{Findings:} The patient is status post median sternotomy and CABG. \hlred{The heart size is enlarged.} The mediastinal and hilar contours are unremarkable. \hlred{Right calcified pleural plaques are seen diffusely} which limits assessment of the underlying pulmonary parenchyma. No focal consolidation, pleural effusion or pneumothorax is clearly demonstrated. There is a suspected left clavicle fracture. \vspace{+5pt}

\textbf{Impression:} \hlred{Right calcified pleural plaques} indicative of prior asbestos exposure. \hlred{There is a moderate left pleural effusion}. No definite acute cardiopulmonary abnormality otherwise noted.
 & `change location', `false prediction', `add contradiction' \\
\hline
\textbf{Findings:} Single frontal view of the chest provided.  There is \hlgreen{no} focal consolidation, effusion, or pneumothorax. The cardiomediastinal silhouette is normal.  Again seen are multiple clips projecting over the left breast and remote left-sided rib fractures.  \hlgreen{No free air below the right hemidiaphragm is seen.} \vspace{+5pt}

\textbf{Impression:} No acute intrathoracic process. & \textbf{Findings:} Single frontal view of the chest provided.  There is \hlred{know} focal consolidation, effusion, or pneumothorax. The cardiomediastinal silhouette is normal.  Again seen are multiple clips projecting over the left breast and remote left-sided rib fractures.  \hlred{There is an ET tube present in the trachea.} No free air below the right hemidiaphragm is seen. \hlred{No free air below the right hemidiaphragm is seen.} \vspace{+5pt}

\textbf{Impression:} No acute intrathoracic process. & `add repetitions', `add medical devices', `change to homophone' \\
\hline
\textbf{Findings:} There is mild-to-moderate cardiomegaly, not significantly changed compared with prior study. There is no pneumothorax. A newly placed endotracheal tube ends \hlgreen{4.3 cm} above the carina. An NG tube is seen ending in the stomach with its tip and side ports beyond the margin of imaging. \vspace{+5pt}

\textbf{Impression:} 1. \hlgreen{Severe acute pulmonary edema.} 2.  \hlgreen{Endotracheal} tube ending 4.3 cm above the carina. & \textbf{Findings:} There is mild-to-moderate cardiomegaly, not significantly changed compared with prior study. There is no pneumothorax. A newly placed endotracheal tube ends \hlred{4.3 mm} above the carina. An NG tube is seen ending in the stomach with its tip and side ports beyond the margin of imaging. \vspace{+5pt}

\textbf{Impression:} 1. \hlred{No pulmonary edema.} 2.  \hlred{Endotrakheal} tube ending 4.3 cm above the carina. & `change measurement', `false negation', `add typo' \\
\hline
\end{tabular}
}
}
}
\end{table}

\subsection{Sentence Level Error Generation Process}
Once the error reports were generated, we used Llama 3.1\cite{meta_ai_team_introducing_nodate} to identify the type of error that was associated within each sentence and at the same time screen for priors. Llama was prompted to generate a python dictionary with each key being an index of the error report and each value containing a list of the label, error class, and original sentence index. Here, label refers to whether the sentence is correct (0), contains an error (1), or is neutral (2), while error class contains the specific category of error introduced if applicable. The original sentence index points back to the index of the original report that was changed to create the sentence in. In the cases where a new sentence was added, the original sentence index was left blank as there was no original sentence used as a reference.

Each error report was then spliced into individual sentences, using the mapping dictionary to determine the original sentence as well as the specific error introduced. Through this methodology, we are able to provide side-by-side comparisons between individual sentences and their associated error sentences. The order of sentences within the original report is maintained, including the position of particular added error sentences. The sentences were manually reviewed to ensure the accuracy of the sentence splicing.

\begin{table}[t!]
\tbl{Examples of ground truth and error sentences generated through the ReXErr sentence splicing and labeling pipeline.}{%
\arrayrulecolor[HTML]{AAACED}
\resizebox{\textwidth}{!}{%
\fontsize{8}{10}\selectfont
{\fontfamily{pcr}
\begin{tabular}{
  >{\raggedright\arraybackslash}p{4.8cm}
  >{\raggedright\arraybackslash}p{4.8cm}
  c
  >{\centering\arraybackslash}p{2.5cm} 
  c
}
\hline
\textbf{Original Sentence} & \textbf{Error Sentence} & \textbf{Label} & \textbf{Error Class} & \textbf{Index} \\
\hline
Findings: Comparison is made to previous study from \_\_\_. & Findings: Comparison is made to previous study from \_\_\_. & 2 & Not Applicable & 0 \\
\hline
There is a right-sided PICC line with distal lead tip at the \hlgreen{cavoatrial junction}. & There is a right-sided PICC line with distal lead tip at the \hlred{mid SVC}. & 1 & Change Position of Device & 1 \\
\hline
There has been removal of the right-sided chest tube. & There has been removal of the right-sided chest tube. & 0 & Not Applicable & 2 \\
\hline
There remains a curvilinear tubular device projecting over the mediastinum. & There remains a curvilinear tubular device projecting over the mediastinum. & 0 & Not Applicable & 3 \\
\hline
This has been seen on \hlgreen{multiple images}. & This has been seen on \hlred{muitiple} images. & 1 & Add Typo & 4 \\
\hline
There is persistent opacity at the left mid lung field and left-sided pleural effusion \hlgreen{which is stable}. & There is persistent opacity at the left mid lung field and left-sided pleural effusion \hlred{which stable}. & 1 & Add Typo & 5 \\
\hline
There is no pulmonary edema. & There is no pulmonary edema. & 0 & Not Applicable & 6 \\
\hline
The right lung is relatively clear. & The right lung is relatively clear. & 0 & Not Applicable & 7 \\
\hline
 & The patient has had placement of an endotracheal tube. & 1 & Add Medical Device & 8 \\
\hline
\end{tabular}
}
}}
\end{table}

\subsection{Validating Error Injection Pipeline}
In order to validate the quality and efficacy of our error injection pipeline, we analyze the projected frequency of every single error category injected across the MIMIC train, dev, and test subsets. Furthermore, a clinician reviewed 100 paired original and error-injected reports to determine the fraction of error reports which are plausible AI-generated or human-written reports. This was done to determine whether the synthesized error reports contain language atypical to radiology reports or very obvious modifications and statements that are not medically plausible which might limit the utility of the synthetic error reports.

\section{Results}
\subsection{Strengths and Limitations of ReXErr}
The ReXErr pipeline was found to proficiently generate errors across all of the error categories listed for the majority of radiology report inputs. It is able to create multiple types of errors in the same report, with variation within each error subtype as well. These errors closely mimic those found in real-world report generation scenarios. Table 3 includes three examples of error reports generated using our report-level error injection pipeline, while Table 4 presents several examples of the sentence-level error generation process, along with the error labeling scheme. Despite ReXErr’s ability to generate errors within the findings and impressions sections, there are still limitations in its ability to maintain consistency in the error injections across both sections. For example, while the first example in Table 3 is handled well, others such as the measurement change in the third example show discrepancies.

\subsection{Consistency Across Error Types}
ReXErr also demonstrates reasonable consistency in distribution of errors inserted across the MIMIC train, dev, and test sets. Certain errors, including “change measurement”, “change name of device”, and “change position of device” are injected less frequently in the dataset due to their reduced prevalence in the original reports. While the weighting mechanism used during sampling helped augment this discrepancy, this quantitative analysis highlights key areas for targeted improvements in developing more robust error injection and correction methods. Table 5 outlines the frequencies of each error type across the train, dev, and test sets, with each value representing the percentage of reports within the given set containing that specific error. Notably, these percentages are relatively consistent across the three different splits. 

\begin{table}[t!]
\tbl{Distribution of errors inserted across the MIMIC train, dev, and test sets using the ReXErr methodology.}{%
\resizebox{0.6\textwidth}{!}{%
\arrayrulecolor[HTML]{AAACED}
\fontsize{8}{10}\selectfont
{\fontfamily{pcr}
\begin{tabular}{lccc}
\hline
Error Category & Train (\%) & Dev (\%) & Test (\%) \\
\hline
Add Medical Device & 33.33 & 33.32 & 33.33 \\
Change Name of Device & 13.64 & 13.47 & 18.91 \\
Change Position of Device & 13.64 & 13.47 & 18.91 \\
Change Severity & 28.71 & 29.88 & 30.18 \\
Change Location & 38.07 & 37.07 & 23.26 \\
False Prediction & 33.33 & 33.32 & 33.33 \\
False Negation & 33.33 & 33.32 & 33.33 \\
Change Measurement & 5.93 & 6.10 & 7.01 \\
Add Opposite Sentence & 25.00 & 24.97 & 24.99 \\
Add Repetitions & 25.00 & 24.97 & 24.99 \\
Change to Homophone & 25.00 & 24.97 & 24.99 \\
Add Typo & 25.00 & 24.97 & 24.99 \\
\hline
\end{tabular}
}}}
\caption{}
\label{aba:fig1}
\end{table}

\subsection{Plausibility of Errors}
Lastly, ReXErr was found to predominantly inject plausible errors within reports. In the sample of 100 ground truth and error-injected reports reviewed by a clinician, 83 of the modified reports were found to be plausible, while only 17 contained errors that were implausible in AI-generated or human reports.

\section{Conclusion}
Synthesizing accurate radiology reports is both difficult and time consuming, even for medical professionals. While automated AI generation approaches are promising in alleviating this workload and more efficiently generating comprehensive reports, they are liable to frequent errors across report content, linguistics, and consistency. Throughout this paper, we present the novel ReXErr method for generating annotated errors on both a report and sentence level. Developed with radiologists, ReXErr captures common AI and human errors in a representative and plausible manner, therefore offering a promising avenue for the development of report screening and correction algorithms as well as improving the accuracy of existing report generation approaches.

\section*{Acknowledgments}

We would like to thank Dr. John Farner and Dr. Rohit Reddy for their valuable clinical input into the error categories and prompts chosen.

\bibliographystyle{ws-procs11x85}
\bibliography{PSB}

\section{Appendix}

\textbf{Appendix A: Base prompt used to inject sampled errors using GPT-4o}

Note: the errors[0], errors[1], and errors[2] present within the prompt below refer to the individual prompts shown in Appendix B for each of the errors that were sampled.
\\ \\
``You will be given a radiology report of a chest X-ray. Your task is to change the statements in the report so that the report is still clinically plausible but has a different meaning than the previous report. You will be given three classes of errors to generate at a time. Make sure to add as many errors as possible to the report, to every single sentence if you can. Look at each sentence, and if you can add an error, make sure to add it. Only one error per sentence. There should not be cases where a sentence within a report does not contain an error unless it is impossible to add an error. Each of the three error classes you are considering should be separated by numbers surrounded by \textless\textless\textless\textgreater\textgreater\textgreater. For example, the first error would start with \textless\textless\textless1\textgreater\textgreater\textgreater. Each error may or may not contain examples. Avoid making multiple errors within the same sentence. Certain error classes, when provided, are labeled as “priority errors” in brackets, meaning that if it is really not possible to add all of the three error types provided, then do your best to add at the very least the priority error. Keep in mind that the goal should still be to add an error to every sentence and use all error classes. Here are the error classes: \textless\textless\textless1\textgreater\textgreater\textgreater " + errors[0] + ``\textless\textless\textless2\textgreater\textgreater\textgreater " + errors[1] + ``\textless\textless\textless3\textgreater\textgreater\textgreater " + errors[2] + ``MAKE sure that the “\textless\textless\textless\textgreater\textgreater\textgreater” characters do not show up in your output- these are only provided for your reference to distinguish between errors. Your output should follow exactly the format that is described below. Here are some guidelines to follow when generating the errors. These guidelines may not be relevant for the given class of error you are tasked with generating, but keep them in consideration. Do not combine unrelated findings in the same sentence. Do not reword sentences when the meaning does not change (ex. do not change ‘normal’ to ‘unremarkable’, ‘multiple’ to ‘several’, or ‘abnormality’ to ‘findings’). Do NOT replace one word with another word that has a similar meaning. For example: ‘noticed’ should not be replaced by ‘seen’. Do not change the order of parts of a sentence, when the meaning does not change.
\\ \\
Keep track of the sentence indexes corresponding to the sentences you change in a report.
\\ \\
\textit{This section of the prompt is added to ensure structure in the output. The dictionary generated here was not used for sentence splicing.}
\\
For a given report, return a new report with the errors in every sentence according to the above paragraph, two new lines, and then a Python dictionary in the following format: {error sentence index : label, explanation, original sentence index]}. The report should be in the exact same format as the original input report, except with the changed sentences. The new report should not contain newlines or any spacing differences compared to the original report. Make sure this format is followed exactly, including the spacing. The label is determined by the following:
\\ \\
0: unchanged sentence \\
1: changed sentence
\\ \\
When the label is 1: `explanation' should contain one statement about the error made in the sentence."
\\ \\
\textbf{Appendix B: Complete prompts instructing the model to inject errors for each particular category}
\\ \\
Note: Some errors were marked [priority error] to encourage the model to insert them in more often when provided with the error, as we observed many cases where the model would not insert those errors when provided with them and opt to insert simpler errors. 

\noindent\hrulefill
\\ \\
\underline{Add Medical Device:}
\\ \\
Add the presence of one or more of devices: add sentences that could be part of a radiology report regarding the presence of one or more of devices such as these: pacemaker, central venous line, NG tube, ET tube, ICD. Only add one sentence per report. \\

\noindent\hrulefill
\\ \\
\underline{Change Name of Device:}
\\ \\
Change the name of a medical instrument: if there is a medical device present in the report, change the name of the medical instrument to a different name that is clinically plausible. If there is no medical device present, DO NOT make any changes and RETURN THE ORIGINAL REPORT (that is allowed for this prompt). DO NOT change a term that is not a medical device to a medical device. DO NOT add any new sentences with medical devices that are not in the original report. Here are some examples:

\bigskip
\noindent \#1 \\
ORIGINAL SENTENCE: Left-sided AICD device is noted with single lead terminating in unchanged position in the right ventricle.\\
OUTPUT: Left-sided dual-chamber pacemaker device is noted with leads terminating in the right atrium and right ventricle.\\
EXPLANATION: Single chamber, AICD not pacemaker

\bigskip
\noindent \#2 \\
ORIGINAL SENTENCE: Left-sided dual-chamber pacemaker device is present with leads terminating in the right atrium and right ventricle.\\
OUTPUT: Left-sided AICD device is noted with leads terminating in the right atrium and right ventricle.\\
EXPLANATION: It is pacemaker, not AICD.
\\ \\
\noindent \#3 \\
ORIGINAL SENTENCE: A right-sided vascular stent is seen within the brachiocephalic vein.\\
OUTPUT: A right-sided PICC line ends in the mid SVC.\\
EXPLANATION: No PICC line, only stent
\\ \\
Here are some examples of where you would not make any changes as there is no presence of a medical device.

\bigskip
\noindent \#1 \\
ORIGINAL SENTENCE: Central vascular engorgement.\\
REASON FOR NO CHANGE: Central vascular engorgement is a finding/condition, not a medical device. Do not turn this into a medical device. 

\bigskip
\noindent \#2 \\
ORIGINAL SENTENCE: Post-operative changes are similar along the right chest wall.\\
REASON FOR NO CHANGE: Post-operative changes are a finding and not a medical device.
\\

\noindent\hrulefill
\\ \\
\underline{Change Position of Device:}
\\ \\
\noindent Change the position of a medical instrument: if there is a medical device location present in the report, change the position of the medical instrument to a different position that is clinically plausible (e.g., ET tube is 5 cm from the carina can be changed to ET tube is 1 cm from the carina). If there is no medical device position, DO NOT make any changes. DO NOT add any positions or change a term that is not a position to a position. Here are some examples of original sentences, the associated error sentence generated, and an explanation for the error.

\bigskip
\noindent \#1 \\
ORIGINAL SENTENCE: Endotracheal tube is placed with its tip located approximately 4.9 cm above the carina.\\
OUTPUT: Endotracheal tube is placed with its tip located approximately 3.5 cm above the carina.\\
EXPLANATION: Distance measurement of ET tube is wrong

\bigskip
\noindent \#2 \\
ORIGINAL SENTENCE: Right IJ central venous catheter projects over the right atrium.\\
OUTPUT: Right IJ central venous catheter ends in the mid SVC.\\
EXPLANATION: Right IJ position lower than mentioned

\bigskip
\noindent \#3 \\
ORIGINAL SENTENCE: The ET tube is in appropriate position.\\
OUTPUT: Portable AP chest radiograph demonstrates the ET tube terminating 2.5 cm above the carina.\\
EXPLANATION: ET tube is closer to carina. If ET tube is in appropriate position, distance should be at least 4 cm from carina

\bigskip
\noindent \#4 \\
ORIGINAL SENTENCE: Left central line terminates in the right atrium.\\
OUTPUT: A left-sided Port-A-Cath terminates in the mid SVC.\\
EXPLANATION: Termination of catheter in right atrium not in mid SVC

\bigskip
\noindent \#5 \\
ORIGINAL SENTENCE: Right PICC terminates near the right subclavian and internal jugular vein confluence with its tip pointing slightly superiorly in the direction of internal jugular vein.\\
OUTPUT: Right PICC line terminates in the mid SVC.\\
EXPLANATION: Right PICC termination is wrong
\\ \\
Here are some examples of where you would not make any changes as there is no presence of a medical device.

\bigskip
\noindent \#1 \\
ORIGINAL SENTENCE: Lateral view shows atherosclerotic coronary calcification in the left circumflex.\\
REASON FOR NO CHANGE: Atherosclerotic coronary calcification is a condition and does not indicate the presence of a device.

\bigskip
\noindent \#2 \\
ORIGINAL SENTENCE: The patient is status post right lower lobectomy.  \\
REASON FOR NO CHANGE: This sentence only indicates the patient’s status and does not specify any device.
 \\

\noindent\hrulefill
\\ \\
\underline{Change Severity:}
\\ \\
\noindent Change severity: change the severity of a finding in the report in a manner that makes clinical sense (e.g., change `mild' to `moderate'; always modifying a sentence). If there is no severity term, DO NOT make any changes, and RETURN THE ORIGINAL REPORT (allowed for this prompt).

\bigskip
\noindent \#1 \\
ORIGINAL SENTENCE: Moderate pulmonary edema.\\
OUTPUT: Mild pulmonary edema.\\
EXPLANATION: Moderate edema is correct so the error report has changed to mild edema.

\bigskip
\noindent \#2 \\
ORIGINAL SENTENCE: The heart is mildly enlarged.\\
OUTPUT: The heart is severely enlarged.\\
EXPLANATION: Mildly enlarged is correct so the error report has changed to severely enlarged.

\bigskip
\noindent \#3 \\
ORIGINAL SENTENCE: There is mild vascular engorgement.\\
OUTPUT: There is moderate vascular engorgement.\\
EXPLANATION: Mildly engorgement is correct so the error report has changed to moderate.

\bigskip
\noindent \#4 \\
ORIGINAL SENTENCE: Small left-sided pleural effusion\\
OUTPUT: Large left-sided pleural effusion\\
EXPLANATION: Small pleural effusion changed to large.
\\ \\
Here are some examples of where you would not make any changes as there is no presence of severity.

\bigskip
\noindent \#1 \\
ORIGINAL SENTENCE: No acute cardiopulmonary abnormality.\\
REASON FOR NO CHANGE: Nothing is changed because changing from “no” to “mild” or “moderate” will create a new finding rather than change the severity of an existing finding.

\bigskip
\noindent \#2 \\
ORIGINAL SENTENCE: No pulmonary edema.\\
REASON FOR NO CHANGE: DO NOT change the phrase “no pulmonary edema” to “mild pulmonary edema” as this creates a finding and is not changing the severity of an existing finding.
 \\

\noindent\hrulefill
\\ \\
\underline{Change Location:}
\\ \\
Change location: change the location or anatomy of a finding in the report in a manner that is still clinically accurate (e.g., change `right' to `left' or `lateral' to `medial', etc; always modifying a sentence). If there is no location term, DO NOT make any changes and DO NOT add a location. If the sentence describes a negative finding (e.g. the lack of a condition or finding) then DO NOT make any changes. For example, if the sentence states “there is no…”, or a phrase with an equivalent meaning, do not make any changes to the sentence. \\

\noindent\hrulefill
\\ \\
\underline{False Prediction:}
\\ \\
\noindent [priority error] Add false predictions: add a finding that is not present in the report (either adding a sentence or modifying a sentence to insert). These should be plausible findings that could be present in the report. Here are some examples:

\bigskip
\noindent \#1 \\
ORIGINAL REPORT: Findings: Frontal and lateral views of the chest are obtained. No focal consolidation, pleural effusion, or evidence of pneumothorax is seen. The cardiac and mediastinal silhouettes are stable. Impression: No acute cardiopulmonary process. No significant interval change. \\
OUTPUT: Findings: Frontal and lateral views of the chest were obtained. There is minimal left base atelectasis. No focal consolidation, pleural effusion, or evidence of pneumothorax is seen. The cardiac and mediastinal silhouettes are stable. There is no overt pulmonary edema. Impression: No acute cardiopulmonary process. \\
EXPLANATION: Added minimal basal atelectasis

\bigskip
\noindent \#2 \\
ORIGINAL REPORT: Findings: PA and lateral views of the chest. There are multiple bilateral rib fractures of varying age as well as old left clavicular fracture. Large hiatal hernia. A heterogeneous opacity concerning for pneumonia is seen in the inferolateral right upper lobe. The left lung is clear. There is no pleural effusion. No pneumothorax. There is no pulmonary vascular congestion. The cardiac, mediastinal, and hilar contours are normal. Impression: 1. Right upper lobe pneumonia. 2. Multiple rib fractures of varying age. Old left clavicular fracture. 3. Stable large hiatal hernia. These findings were discussed with Dr. \_ at 3:30 p.m. on \_ by telephone. \\
OUTPUT: Findings: PA and lateral views of the chest. There is a large hiatal hernia. The lungs are clear. There is no pleural effusion or pneumothorax. The cardiac, mediastinal, and hilar contours are normal. There is a large osteolytic lesion in the right posterior rib. There is a small compression deformity in the lower thoracic spine. Impression: 1. No evidence of pneumonia. 2. Large hiatal hernia. 3. Stable moderate cardiomegaly. \\
EXPLANATION: Missed lung opacity, gave it as large rib lesion. Added compression deformity.

\bigskip
\noindent \#3 \\
ORIGINAL REPORT: Findings: AP single view of the chest has been obtained with patient in upright position. There is no evidence of pneumothorax in the apical area on either left or right side. In comparison with the next preceding chest examination of \_, at that time described pulmonary abnormalities including a left lower lobe mass persists. Impression: No evidence of pneumothorax following transbronchial biopsy intervention. \\
OUTPUT: Findings: AP single view of the chest has been obtained with patient in semi-upright position. Comparison is made with the next preceding similar study of \_. The patient is now status post biopsy of a left-sided mass in the left lower lobe area. No pneumothorax is seen. No new pulmonary abnormalities are identified. The previously described right-sided apical mass is again seen. Impression: No evidence of pneumothorax following biopsy procedure. \\
EXPLANATION: Added right-sided mass.

\bigskip
\noindent \#4 \\
ORIGINAL REPORT: Findings: There is no focal consolidation, pleural effusion or pneumothorax. Bilateral nodular opacities that most likely represent nipple shadows. The cardiomediastinal silhouette is normal. Clips project over the left lung, potentially within the breast. The imaged upper abdomen is unremarkable. Chronic deformity of the posterior left sixth and seventh ribs are noted. Impression: No acute cardiopulmonary process. \\
OUTPUT: Findings: There is no focal consolidation, or pneumothorax. Bilateral nodular opacities that most likely represent nipple shadows. There is mild cardiomegaly as well as trace bilateral effusions within the fissures. Clips project over the left lung, potentially within the breast. The imaged upper abdomen is unremarkable. Chronic deformity of the posterior left sixth and seventh ribs are noted. Impression: No acute cardiopulmonary process. \\
EXPLANATION: Changed cardiomediastinal silhouette is normal to mild cardiomegaly with effusions.

\bigskip
\noindent \#5 \\
ORIGINAL REPORT: Findings: The cardiac, mediastinal and hilar contours are normal. Pulmonary vasculature is normal. Lungs are clear. No pleural effusion or pneumothorax is present. Multiple clips are again seen projecting over the left breast. Remote left-sided rib fractures are also re-demonstrated. Impression: No acute cardiopulmonary abnormality. \\
OUTPUT: Findings: The cardiac, mediastinal and hilar contours are normal. Pulmonary vasculature is normal. Lungs are clear. No pneumothorax is present. A small left pleural effusion persists as well as mild pulmonary vascular congestion. Multiple clips are again seen projecting over the left breast. Remote left-sided rib fractures are also re-demonstrated. Impression: No acute cardiopulmonary abnormality. \\
EXPLANATION: Added a small left pleural effusion and pulmonary congestion.

\bigskip
\noindent \#6 \\
ORIGINAL REPORT: Findings: Single frontal view of the chest provided. There is no focal consolidation, effusion, or pneumothorax. The cardiomediastinal silhouette is normal. Again seen are multiple clips projecting over the left breast and remote left-sided rib fractures. No free air below the right hemidiaphragm is seen. Impression: No acute intrathoracic process. \\
OUTPUT: Findings: Single frontal view of the chest provided. There is no focal consolidation, effusion, or pneumothorax. The cardiomediastinal silhouette is normal. Again seen are multiple clips projecting over the left breast and remote left-sided rib fractures. Bibasilar patchy opacities could reflect aspiration, atelectasis or infection. No free air below the right hemidiaphragm is seen. Impression: No acute intrathoracic process. \\
EXPLANATION: Added bibasilar patchy opacities could reflect aspiration, atelectasis or infection.

\bigskip
\noindent \#7 \\
ORIGINAL REPORT: Findings: The lungs are clear of focal consolidation, pleural effusion or pneumothorax. The heart size is normal. The mediastinal contours are normal. Multiple surgical clips project over the left breast, and old left rib fractures are noted. Impression: No acute cardiopulmonary process. \\
OUTPUT: Findings: The lungs are clear of focal consolidation, pleural effusion or pneumothorax. The heart size is normal. The mediastinal contours are normal. Multiple surgical clips project over the left breast, and old left rib fractures are noted. Orthopedic hardware seen in the right humeral head. Impression: No acute cardiopulmonary process. \\
EXPLANATION: Added orthopedic hardware in the right humeral head.

\bigskip
\noindent \#8 \\
ORIGINAL REPORT: Findings: PA and lateral views of the chest provided. Lung volumes are somewhat low. There is no focal consolidation, effusion, or pneumothorax. The cardiomediastinal silhouette is normal. Imaged osseous structures are intact. No free air below the right hemidiaphragm is seen. Impression: No acute intrathoracic process. \\
OUTPUT: Findings: PA and lateral views of the chest provided. Lung volumes are somewhat low. The lungs are hyperinflated. There is no focal consolidation, effusion, or pneumothorax. The cardiomediastinal silhouette is normal. Imaged osseous structures are intact. No free air below the right hemidiaphragm is seen. Impression: No acute intrathoracic process. \\
EXPLANATION: Added hyperinflation to the lungs.

\bigskip
\noindent \#9 \\
ORIGINAL REPORT: Findings: As compared to the prior examination dated \_, there has been no significant interval change. There is no evidence of focal consolidation, pleural effusion, pneumothorax, or frank pulmonary edema. The cardiomediastinal silhouette is within normal limits. There is persistent thoracic kyphosis with mild wedging of a mid thoracic vertebral body. Impression: No evidence of acute cardiopulmonary process. \\
OUTPUT: Findings: As compared to the prior examination dated \_, there has been no significant interval change. Hyperexpansion consistent with COPD. There is no evidence of focal consolidation, pleural effusion, pneumothorax, or frank pulmonary edema. The cardiomediastinal silhouette is within normal limits. There is persistent thoracic kyphosis with mild wedging of a mid thoracic vertebral body. Impression: No evidence of acute cardiopulmonary process. \\
EXPLANATION: Added hyperexpansion consistent with COPD.

\bigskip
\noindent \#10 \\
ORIGINAL REPORT: Findings: Heart size is normal. Mediastinal contours are normal with mild aortic tortuosity. Post-surgical changes in the right hemithorax are stable including thickening of the pleura along the costal surface and blunting of the costophrenic sulcus. The right sixth rib surgical fracture is redemonstrated. There are no new lung nodules identified. Impression: Stable chest radiograph. \\
OUTPUT: Findings: Heart size is normal. Mediastinal contours are normal with mild aortic tortuosity. Post-surgical changes in the right hemithorax are stable including thickening of the pleura along the costal surface and blunting of the costophrenic sulcus. Chronic interstitial process with superimposed airspace opacities likely representing pneumonia. The right sixth rib surgical fracture is redemonstrated. There are no new lung nodules identified. Impression: Airspace opacities likely representing pneumonia. \\
EXPLANATION: Added the presence of chronic interstitial process likely representing pneumonia.

\bigskip
\noindent \#11 \\
ORIGINAL REPORT: Findings: Mild to moderate enlargement of the cardiac silhouette is unchanged. The aorta is calcified and diffusely tortuous. The mediastinal and hilar contours are otherwise similar in appearance. There is minimal upper zone vascular redistribution without overt pulmonary edema. No focal consolidation, pleural effusion or pneumothorax is present. The osseous structures are diffusely demineralized. Impression: No radiographic evidence for pneumonia. \\
OUTPUT: Findings: Status post prior median sternotomy. Mild to moderate enlargement of the cardiac silhouette is unchanged. The aorta is calcified and diffusely tortuous. The mediastinal and hilar contours are otherwise similar in appearance. There is minimal upper zone vascular redistribution without overt pulmonary edema. No focal consolidation, pleural effusion or pneumothorax is present. The osseous structures are diffusely demineralized. Impression: No radiographic evidence for pneumonia. \\
EXPLANATION: Added status post prior median sternotomy when it’s not.

\bigskip
\noindent \#12 \\
ORIGINAL REPORT: Findings: The lungs are clear of consolidation, effusion, or edema. Cardiac silhouette is top normal. Descending thoracic aorta is tortuous with atherosclerotic calcification seen at the arch. No acute osseous abnormalities identified. Impression: No acute cardiopulmonary process. \\
OUTPUT: Findings: The lungs are clear of consolidation, effusion, or edema. Streaky opacities in the lung bases likely reflect atelectasis. Cardiac silhouette is top normal. Descending thoracic aorta is tortuous with atherosclerotic calcification seen at the arch. No acute osseous abnormalities identified. Impression: No acute cardiopulmonary process. \\
EXPLANATION: Added streaky opacities in the lung that reflect atelectasis.
 \\

\noindent\hrulefill
\\ \\
\underline{False Negation:}
\\ \\
\noindent [priority error] Change a particular finding from the report from present to absent by changing a sentence to indicate absence of the positive clinical finding. When you change the finding, remove all details associated with the original finding: for example severity, location, and additional details that would only be relevant if the finding existed. Only change POSITIVE findings (i.e., you can remove `cardiomegaly' but not `the heart is normal in size'). Do not change NEGATIVE findings to positive findings. DO NOT REMOVE/OMIT SENTENCES FROM THE ORIGINAL REPORT. Negative findings include heart is normal, lungs are clear, cardiac/mediastinal/hilar contours are normal. Here are some examples:
\\ \\
Finding: “Clips project over the left lung, potentially within the breast.” \\
What not to do: “Clips do not project over the left lung, potentially within the breast.” \\
What you should do: “No clips seen” as this removes all details (where the clips were seen). \\
Explanation: Clips projecting over a lung is a positive clinical finding, so you can change this sentence. Leaving in the details, such as the left lung and the breast is not relevant when there is no finding, hence they should not be included in the negation.
\\ \\
Finding: “No pleural effusion or pneumothorax is seen.” \\
What not to do: “There is a right pleural effusion.” \\
Explanation: Saying that there is no pleural effusion or pneumothorax is NOT a positive finding, so you should not change this sentence
\\ \\
Finding: “Bilateral nodular opacities that most likely represent nipple shadows” \\
What not to do: “No bilateral nodular opacities that most likely represent nipple shadows.” \\
What you should do: “No nodular opacities” as this also removes location. \\
Explanation: This is a positive finding, so you can change it. ``bilateral nodular opacities" and ``nipple shadows" are both not relevant at all when there is no finding, meaning that they should be removed in the negation.
\\ \\
Finding: ``There are innumerable bilateral scattered small pulmonary nodules which are better demonstrated on recent CT." \\
What not to do: ``There are no discernible bilateral scattered small pulmonary nodules." \\
What you should do: “No pulmonary nodules.” \\
Explanation: The presence of nodules is a positive findings, but their location is irrelevant when nodules are not present so they should not be mentioned in the negation.
\\ \\
Finding: ``Otherwise, the lungs are clear." \\
What not to do not do: ``No abnormal lung findings are noted." \\
Explanation: Saying that the lungs are clear is NOT a positive finding, so you should not change this sentence
\\ \\
Finding: ``Findings: There is mild pulmonary edema with superimposed region of more confluent consolidation in the left upper lung." \\
What not to do: ``Findings: There is mild pulmonary edema with no regions of more confluent consolidation in the lung." \\
What you should do: “There is mild pulmonary edema with no consolidation.” \\
Explanation: Pulmonary edema is a positive finding so you may change the sentence. However, specifying the areas of confluent consolidation is irrelevant when the finding is negated, so it should not be included.
\\ \\
Finding: ``Descending thoracic aorta is tortuous with atherosclerotic calcification seen at the arch." \\
What not to do: ``Descending thoracic aorta is not tortuous with no evidence of atherosclerotic calcification." \\
What you should do: “Aorta appears normal.” \\
Explanation: The aorta being tortuous with atherosclerotic calcification is a positive medical finding, so you can change the sentence. However, the location of the calcification as well as the area of the aorta is not relevant if the aorta is normal, so this information should be left out on the negation.
\\ \\
Finding: ``Impression: New moderate left pleural effusion with adjacent atelectasis in the left lung base." \\
What not to do: ``Impression: New moderate left pleural effusion with adjacent atelectasis in the left lung base. No pleural effusion." \\
What you should do: “No pleural effusion.” \\
Explanation: Make sure that you are not adding or removing any sentences. For example, to negate this statement, you should not add another sentence saying ``no pleural effusion", as this would be a contradiction, a different error type. Instead, the starting sentence should be entirely replaced with ``no pleural effusion".
\\ \\
Finding: ``The right middle lobe pneumonia seen on recent CT is not clearly differentiated, but the right heart border is obscured." \\
What not to do: ``No right middle lobe pneumonia is seen on recent CT." \\
What you should do: “No right middle lobe pneumonia is seen on recent CT.” \\
Explanation: The presence of pneumonia is a positive medical finding, so you may change this sentence. You should not include ``right middle lobe" in the negated sentence, as it does not make sense when there is no finding of pneumonia.
\\ \\
Finding: ``Heart size top-normal." \\
What not to do: ``No changes in heart size" \\
Explanation: The heart size being top normal is not a positive medical finding (it indicates the heart is normal), so no changes should be made to the sentence.
\\ \\
As shown in all of these examples, changing an error from present to absent is more than just changing a single word, or adding a ``not". Think about how the radiology report would be written if the positive medical finding in question was not actually present when re-writing the sentence as a negation.
 \\

\noindent\hrulefill
\\ \\
\underline{Change Measurement:}
\\ \\
\noindent Change measurement units or values: if there is a measurement for a device/finding present, change the units of measurement (e.g., change `cm' to `mm' or `mm' to `cm', etc; DO NOT change to inches or feet, keep units within the metric system) or change the value of the measurement to a different but still reasonable value (e.g. change `4.9 cm' to `5.8 cm'). If there is no measurement, DO NOT make any changes and RETURN THE ORIGINAL REPORT. DO NOT ADD ANY MEASUREMENT PHRASES. DO NOT change the units for any times of day.

\bigskip
\noindent \#1 \\
ORIGINAL SENTENCE: Endotracheal tube is placed with its tip located approximately 4.9 cm above the carina. \\
OUTPUT: Endotracheal tube is placed with its tip located approximately 5.8 cm above the carina. \\
EXPLANATION: Measurement value changed from 4.9 cm to 5.8 cm

\bigskip
\noindent \#2 \\
ORIGINAL SENTENCE: To achieve a better placement, the tube needs to be retracted by approximately 10 cm and then it may be redirected toward the lower stomach. \\
OUTPUT: To achieve a better placement, the tube needs to be retracted by approximately 10 mm and then it may be redirected toward the lower stomach. \\
EXPLANATION: measurement changed from cm to mm

\bigskip
\noindent \#3 \\
ORIGINAL SENTENCE: 4 mm nodule projecting over the retrosternal clear space on the lateral view appears new compared with chest x-rays from \_\_\_ and which may be followed up with a chest CT. \\
OUTPUT: 4 cm nodule projecting over the retrosternal clear space on the lateral view appears new compared with chest x-rays from \_\_\_ and which may be followed up with a chest CT.  \\
EXPLANATION: measurement changed from mm to cm
\\ \\
Here are some examples of where you would not make any changes as there is no presence of any measurement unit:

\bigskip
\noindent \#1 \\
ORIGINAL SENTENCE: Enteric tube courses below the level of the diaphragm. \\
REASON FOR NO CHANGE: There is no measurement unit (mm, cm, inch, km, etc)

\bigskip
\noindent \#2 \\
ORIGINAL SENTENCE: Aorta is markedly tortuous, unchanged. \\
REASON FOR NO CHANGE: DO NOT change ``Aorta is markedly tortuous, unchanged.” to ``Aorta is markedly tortuous and measures about 2 inches, unchanged” because you are adding the phrase “measures about 2 inches” when there originally was no such phrase or measurement.
 \\

\noindent DO NOT change both the measurement unit and the measurement value at the same time. For example, here are some incorrect changes you should not make:

\bigskip
\noindent \#1 \\
Finding: ``Endotracheal tube terminates 4.9 mm above the carina." \\
What not to do: ``Endotracheal tube terminates 6.7 cm above the carina." \\
What you should do: ``Endotracheal tube terminates 4.9 cm above the carina." OR ``Endotracheal tube terminates 6.7 mm above the carina." \\
Explanation: You should only change either the units (ex. 4.9 mm to 4.9 cm) or the value of the measure (ex. 4.9 mm to 6.7 mm)

\bigskip
\noindent \#2 \\
Finding: ``There is a subtle nodular opacity projecting over the left lower chest measuring approximately 1 mm." \\
What not to do: ``There is a subtle nodular opacity projecting over the left lower chest measuring approximately 2 cm." \\
What you should do: ``There is a subtle nodular opacity projecting over the left lower chest measuring approximately 1 cm." OR ``There is a subtle nodular opacity projecting over the left lower chest measuring approximately 2 mm" \\
Explanation: You should only change either the units (ex. 1 mm to 1 cm) or the value of the measure (ex. 1 mm to 2 mm) \\
 

\noindent\hrulefill
\\ \\
\underline{Add Contradictions:}
\\ \\
\noindent [priority error] Add an opposite sentence: Add/alter a statement that is the opposite of another statement earlier in the same report. Ensure that the opposites are clear and evident either between the findings and impressions sections of the report. The added sentence should demonstrate a clear inconsistency in the medical assessment and between NEW findings (not findings from prior reports). DO NOT ADD statements by themselves that don’t oppose anything. Below are detailed examples to illustrate:

\bigskip
\noindent \#1: \\
ORIGINAL REPORT: No pleural effusion or pneumothorax is seen. Impression: No acute cardiopulmonary process. \\
ERROR REPORT (what you should output): No pleural effusion or pneumothorax is seen. Impression: Mild pulmonary edema and small bilateral pleural effusions. \\
EXPLANATION: Findings in output say no pleural effusion and small pleural effusion so there is an opposite injected.

\bigskip
\noindent \#2: \\
ORIGINAL REPORT: Findings: Diffuse bilateral mainly basilar parenchymal opacities consistent with moderate pulmonary edema. Impression: Moderate pulmonary edema and small bilateral pleural effusions and cardiomegaly consistent with congestive heart failure. \\
ERROR REPORT (what you should output): Findings: There are bilateral diffuse interstitial opacities with a perihilar predominance, consistent with moderate pulmonary edema.  There are small bilateral pleural effusions. Impression: 1.  Mild pulmonary edema. 2.  Small bilateral pleural effusions. \\
EXPLANATION: Findings in output say moderate pulmonary edema and mild pulmonary edema so there is a conflict between moderate and mild.

\bigskip
\noindent \#3: \\
ORIGINAL REPORT: Findings: Single frontal radiograph of the chest was performed and reveals no acute cardiopulmonary process. Impression: No acute cardiopulmonary process. \\
ERROR REPORT (what you should output): Findings: There is increased opacity in the left upper lobe and right upper lobe, which may represent multifocal pneumonia. Impression: No acute cardiopulmonary process. \\
EXPLANATION: multifocal pneumonia and no acute cardiopulmonary process indicate opposition in output report.

\bigskip
\noindent \#4: \\
ORIGINAL REPORT: Findings: Right IJ central venous catheter projects over the right atrium. Impression: 1. Interval placement of a right IJ catheter with tip in the right atrium, consider pulling back by 3 cm for optimal placement. \\
ERROR REPORT (what you should output): Findings: Right IJ central venous catheter ends in the mid SVC. Impression: 1. Right IJ central venous catheter terminates in the right atrium. \\
EXPLANATION: Right IJ position ending in mid SVC oppose position terminating in right atrium in output report.

\bigskip
\noindent Note that in all of these examples, the report was changed so that now in the error report, there are two separate sentences that contradict each other. It is not enough to simply negate the finding within a single sentence. There should be two separate sentences in the final error report which directly contradict one another.
 \\

\noindent\hrulefill
\\ \\
\underline{Add Repetitions:}
\\ \\
Add repetitions of sentences. Make sure to mark the second instance of the sentence as the error, not the first. If the report contains two repeated sentences with some sentences in between, mark the sentence that appears later within the report as the error. For example if the report contains the following sentences: “There is no pleural effusion or pneumothorax. Clips extend over the left lung. There is no pleural effusion or pneumothorax.”, you would only characterize the third sentence as the error, not the first. Have one repetition per report. Make sure that the repeated sentence is written exactly the same as the original sentence and that both sentences show up in the final error report. \\

\noindent\hrulefill
\\ \\
\underline{Change to Homophone:}
\\ \\
Change word to homophones: change a word in the report to a homophone of that word (e.g., replacing the word ‘four’ with ‘for’ in “Right pneumothorax with for mm apical pleural separation”, replacing ‘two’ with ‘to’ or ‘too’, or replacing “psoas” with “so as”, etc). \\

\noindent\hrulefill
\\ \\
\underline{Add Typo:}
\\ \\
Add a typographical error: add a typographical error in the report (e.g., change `right' to `rigth', `no' to `now', etc; always modifying a sentence). Make sure that the typos still contain some resemblance to the original world (for example, do not change “clear” to “learn” as they are completely different words). \\

\noindent\hrulefill
\\ \\
\textbf{Appendix C: Sentence splicing prompt used to prompt Llama 3.1.}
\\ \\
You will be given two reports, an error report and the original report and the lengths of the two reports. If the lengths of the two reports are the same, your job is to match each sentence of the error report to the original report. The sentences will resemble each other with some phrasing changes. Return a dictionary where you have each sentence of the original report matched to each sentence of the error report. Here is an example:
\\ \\
Original Report: Impression: As compared to \_\_\_, the lung volumes have slightly decreased.  Signs of mild overinflation and moderate pleural effusions persist.  Moderate cardiomegaly.  Elongation of the descending aorta.  No pneumonia.
\\ \\
Error Report: Impression: As compared to \_\_\_, the lung volumes have significantly increased. Signs of severe overinflation and minor pleural effusions persist. Mild cardiomegaly. Elongation of the ascending aorta. No pneumonia. \\ \\
Length of Original Report: 5 \\
Length of Error Report: 5
\\ \\
Output: \{`Impression: As compared to \_\_\_, the lung volumes have slightly decreased.' : `Impression: As compared to \_\_\_, the lung volumes have significantly increased.', `Signs of mild overinflation and moderate pleural effusions persist.' : `Signs of severe overinflation and minor pleural effusions persist.', `Moderate cardiomegaly.' : `Mild cardiomegaly.', `Elongation of the descending aorta.' : `Elongation of the ascending aorta.', `No pneumonia.' : `No pneumonia.'\}
\\ \\
If the lengths of the reports are not the same, your job is still to match each sentence of the error report to the original report, however now you will also add entries in the output for ``added sentences" or ``missing sentences". For example, if the error report includes repetitions, mentions of added medical devices, and findings not present in the original report, then you would have a dictionary entry of `' : `added sentence'. If the error report has an omission of something in the original report, you would have a dictionary entry of `omitted sentence' : `'. Here is an example:
\\ \\
Original Report: Findings: NG tube is coiled in the stomach.  Right PICC in lower SVC is unchanged in position.  Cardiac size is normal.  Mild bibasilar opacities consistent with atelectasis, unchanged compared to chest radiograph performed earlier in the same day.  There is no pneumothorax or pleural effusion. Impression: NG tube in expected position with tip coiled in the stomach.  No other interval change since chest radiograph performed earlier on the same day. \\ \\
Error Report: Findings: NG tube is coiled in the upper part of the duodenum. Right PICC in proximal SVC is unchanged in position. Mild bibasilar opacities consistent with atelectasis, unchanged compared to chest radiograph performed earlier in the same day. There is no pneumothorax or pleural effusion. Impression: NG tube in unexpected position with tip coiled in duodenum. Large bilateral pleural effusions are noted. \\ \\
Length of Original Report: 7 \\
Length of Error Report: 6 \\ \\
Output: \{`Findings: NG tube is coiled in the stomach.' : `Findings: NG tube is coiled in the upper part of the duodenum.', `Right PICC in lower SVC is unchanged in position.' : `Right PICC in proximal SVC is unchanged in position.', `Cardiac size is normal.' : `', `Mild bibasilar opacities consistent with atelectasis, unchanged compared to chest radiograph performed earlier in the same day. ' : `Mild bibasilar opacities consistent with atelectasis, unchanged compared to chest radiograph performed earlier in the same day.', `There is no pneumothorax or pleural effusion.' : `There is no pneumothorax or pleural effusion.', `Impression: NG tube in expected position with tip coiled in the stomach.' : `Impression: NG tube in unexpected position with tip coiled in duodenum.', `' : `Large bilateral pleural effusions are noted.', `No other interval change since chest radiograph performed earlier on the same day.' : `'  \}
\\ \\
Do not leave any sentences out from both the error and the original report. Each sentence should have an entry, whether matched to another sentence, or matched to an empty string. Do not add empty entries like `':`'. Return ONLY the dictionary output, with no text before or after.
\\ \\
\textbf{Appendix D: Sentence labeling prompt used to prompt Llama 3.1.}
\\ \\
You will be given a python dictionary mapping the sentences of an original report to the error report \{original sentence : error sentence\}. Your job is to label the error inserted into the sentence of the error report from the original report. Return only a Python dictionary in the following format: \{original sentence : [label, error class, error sentence]\}. This means that the length of the returned dictionary is the same as that of the given dictionary. Make sure this format is followed exactly, including the spacing. The label is determined by the following:
\\ \\
0: unchanged sentence\\
1: changed sentence\\
2: neutral sentence\\
\\
Neutral sentences (label 2) are sentences that offer comparisons, suggestions, or recommendations that are not necessarily correct or incorrect and should therefore be labeled as neutral. If there is any mention of the words ``previous'' or ``prior'' mark the sentence as having a reference to a prior. NOTE: even if there is a change in the sentence, if you believe it falls in the neutral category, mark it in that category. You should also screen for sentences that contain comparisons between current findings and previous findings. These include any sentences that explicitly refer to a different time point (usually using words such as ``since'' or ``compared to''). Also label sentences which contain no clinical findings as neutral. See examples number 5 and 6 below for reference of a sentence without clinical findings. Here are the examples of sentences that you would classify as neutral:
\\ \\
1. ``Unexplained severe rightward deviation of the trachea without tracheal narrowing at the level of the thoracic inlet, not markedly changed since \_\_'': this sentence is neutral because it designates a ``change since'', referencing an earlier time point\\
2. ``Stable COPD'': this sentence is neutral because the word ``stable'' is a comparison to a previous time point.\\
3. ``There is a high level of focal consolidation which has been stable since \_\_'': this sentence is neutral because it indicates ``stable since'' a particular past time point\\
4. ``Compared to chest radiographs since \_\_'': this sentence is neutral because it references prior radiographs\\
5. ``Received note from Dr.\_\_ on \_\_'': this sentence is neutral because it contains no clinical findings.\\
6. ``Dr. \_\_ communicated the above results to Dr. \_\_ at 8:55 am on \_\_ by telephone.''
\\ \\
Within the dictionary, the error class falls into the following 13 error classes. Please choose the one that is most relevant whenever there is a change in the sentence. When inputting them into the dictionary in the error class category, write out the name of the error exactly as it is written here. If there is no error, the label is 13 ``Not applicable.'' Some of the items listed below contain colons followed by explanations, but only use the part before the colon as the label.
\\ \\
1. Add Medical Device\\
2. Change Name of Already Present Device\\
Example: \{ORIGINAL SENTENCE: Left-sided AICD device is noted with single lead terminating in unchanged position in the right ventricle. ERROR SENTENCE: Left-sided dual-chamber pacemaker device is noted with leads terminating in the right atrium and right ventricle. EXPLANATION: Single chamber, AICD not pacemaker\}\\
3. Change Position of Already Present Device\\
Example: \{ORIGINAL SENTENCE: Right IJ central venous catheter projects over the right atrium. ERROR SENTENCE: Right IJ central venous catheter ends in the mid SVC. EXPLANATION: Right IJ position lower than mentioned\}\\
4. Change Already Present Severity: eg. from mild to moderate or severe or vice versa\\
Example: \{ORIGINAL SENTENCE: Moderate pulmonary edema. ERROR SENTENCE: Mild pulmonary edema. EXPLANATION: Moderate edema is correct so the error report has changed to mild edema.\}\\
5. Change Location: e.g., changing `right' to `left' or `lateral' to `medial', etc. (note that this applies for a location change of a finding, separate from changing the position of a medical device, which is a separate category)\\
6. False Negation: removing a positive finding either by rewording or deleting the sentence\\
Example: \{ORIGINAL SENTENCE: Clips project over the left lung, potentially within the breast. ERROR SENTENCE: No clips seen EXPLANATION: Positive finding of clips projecting over the left lung was removed\}\\
7. False Prediction: adding a fake positive finding or changing something from normal to not normal\\
Example: \{ORIGINAL REPORT: Findings: Frontal and lateral views of the chest are obtained. No focal consolidation, pleural effusion, or evidence of pneumothorax is seen. The cardiac and mediastinal silhouettes are stable. Impression: No acute cardiopulmonary process. No significant interval change. ERROR REPORT: Findings: Frontal and lateral views of the chest were obtained. There is minimal left base atelectasis. No focal consolidation, pleural effusion, or evidence of pneumothorax is seen. The cardiac and mediastinal silhouettes are stable. There is no overt pulmonary edema. Impression: No acute cardiopulmonary process. EXPLANATION: Sentence index 1 of the error report adds a false prediction of minimal basal atelectasis\}\\
8. Change measurement: eg. from mm to cm\\
9. Add opposite sentence: contradictory statement added within the same report\\
\{ORIGINAL REPORT: No pleural effusion or pneumothorax is seen. Impression: No acute cardiopulmonary process. ERROR REPORT: No pleural effusion or pneumothorax is seen. Impression: Mild pulmonary edema and small bilateral pleural effusions. EXPLANATION: Findings in error report say no pleural effusion. Sentence index 1 in the error report is a contradiction as it contradicts the finding of no pleural effusion in sentence index 0.\}\\
10. Add repetitions\\
11. Change to Homophone (e.g. ``seen" to ``scene")\\
12. Add typo\\
13. Not applicable– the sentences are the same/very similar
\\ \\
Here is an example:
\\ \\
Dictionary: \{`Findings: The lung volumes are low.' : `Findings: The lung volumse are low.', `The cardiac, mediastinal and hilar contours appear unchanged, allowing for differences in technique.' : `The cardiac, mediastinal and hilar contours appear unchanged, allowing for differences in technique.', `There are a number of round nodular densities projecting over each upper lung, but more numerous and discretely visualized in the left upper lobe, similar to prior study.' : `There are a number of round nodular densities projecting over each lower lung, but more numerous and discretely visualized in the right lower lobe, similar to prior study.', `However, in addition, there is a more hazy widespread opacity projecting over the left mid upper lung which could be compatible with a coinciding pneumonia.' : `However, in addition, there is a more hazy widespread opacity projecting over the left mid upper lung which could be compatible with a coinciding pneumonia.', `Pulmonary nodules in the left upper lobe are also not completely characterized on this study.' : `Pulmonary nodules in the right lower lobe are also not completely characterized on this study.', `There is no pleural effusion or pneumothorax.' : `There is no pleural effusion or pneumothorax.', `Post-operative changes are similar along the right chest wall.' : `Post-operative changes are similar along the left chest wall.', `Impression: Increasing left lung opacification which may reflect pneumonia superimposed on metastatic disease, although other etiologies such as lymphangitic pattern of metastatic spread could be considered.' : `Impression: Increasing right lung opacification which may reflect pneumonia superimposed on metastatic disease, although other etiologies such as lymphangitic pattern of metastatic spread could be considered.', `CT may be helpful to evaluate further if needed clinically.' : `CT may be helpful to evaluate further if needed clinically.'\} \\ \\
Output: \{`Findings: The lung volumes are low.' : [1, 12, `Findings: The lung volumse are low.'], `The cardiac, mediastinal and hilar contours appear unchanged, allowing for differences in technique.' : [0, 13, `The cardiac, mediastinal and hilar contours appear unchanged, allowing for differences in technique.'], `There are a number of round nodular densities projecting over each upper lung, but more numerous and discretely visualized in the left upper lobe, similar to prior study.' : [1, 5, `There are a number of round nodular densities projecting over each lower lung, but more numerous and discretely visualized in the right lower lobe, similar to prior study.'], `However, in addition, there is a more hazy widespread opacity projecting over the left mid upper lung which could be compatible with a coinciding pneumonia.' : [0, 13, `However, in addition, there is a more hazy widespread opacity projecting over the left mid upper lung which could be compatible with a coinciding pneumonia.'], `Pulmonary nodules in the left upper lobe are also not completely characterized on this study.' : [1, 5, `Pulmonary nodules in the right lower lobe are also not completely characterized on this study.'], `There is no pleural effusion or pneumothorax.' : [0, 13, `There is no pleural effusion or pneumothorax.'], `Post-operative changes are similar along the right chest wall.' : [0, 5, `Post-operative changes are similar along the left chest wall.'], `Impression: Increasing left lung opacification which may reflect pneumonia superimposed on metastatic disease, although other etiologies such as lymphangitic pattern of metastatic spread could be considered.' : [0, 5, `Impression: Increasing right lung opacification which may reflect pneumonia superimposed on metastatic disease, although other etiologies such as lymphangitic pattern of metastatic spread could be considered.'], `CT may be helpful to evaluate further if needed clinically.' : [0, 13, `CT may be helpful to evaluate further if needed clinically.']\}
\\ \\
Note that you can have a label of 2 for neutral but a different error class. For example, notice that here there is a reference to a comparison so the label is 2; however there is a change from decreased to increased for the lung volumes. This is a change in severity. Thus the error class label is 4. \{`Impression: As compared to \_\_\_, the lung volumes have slightly decreased.': [2, 4, `Impression: As compared to \_\_\_, the lung volumes have significantly increased.']\}
\\ \\
Note that if there is an entry in the dictionary that is of this format `` : `sentence', that means it is an added sentence. The added sentence errors can either be 1, 7, 9, or 10. If the sentence is a repetition of a previous dictionary entry, then it is a repetition error, so class number 10. If it is a sentence about a medical device, then it is an added device so it is error 1. If it is a new finding, then it is an added predicted finding so error 7. If it contradicts a previous dictionary entry, then it is an added opposite sentence so error 9. If there is an entry in the dictionary that looks like `sentence': `', then the only possible error class is 6.
\\ \\
Return ONLY the output dictionary, with no text before or after.

\end{document}